%% file: main.tex
\documentclass[10pt,twocolumn,letterpaper]{article}

\usepackage[pagenumbers]{cvpr} 

\input{preamble}

\makeatletter
\renewcommand{\paragraph}{%
  \@startsection{paragraph}{4}%
  {\z@}{0.8ex \@plus 0.2ex \@minus .2ex}{-1em}%
  {\normalfont\normalsize\bfseries}%
}
\makeatother
\usepackage{fancyvrb}
\usepackage{calc}

\newlength{\kubric}
\newlength{\kubricb}
\newlength{\kubricc}

\newlength{\ucsd}
\newlength{\ucsdb}

\definecolor{cvprblue}{rgb}{0.21,0.49,0.74}
\usepackage[pagebackref,breaklinks,colorlinks,allcolors=cvprblue]{hyperref}

\def\paperID{4429} 
\def\confName{CVPR}
\def\confYear{2026}

\title{GRVS: a Generalizable and Recurrent Approach to\\ Monocular Dynamic View Synthesis}

\author{Thomas Tanay* \quad Mohammed Brahimi* \quad Michal Nazarczuk \quad Qingwen Zhang \\ Sibi Catley-Chandar \quad Arthur Moreau \quad Zhensong Zhang \quad Eduardo Pérez-Pellitero\vspace{0.2cm}\\
Huawei Noah's Ark Lab}

\begin{document}
\maketitle

\let\thefootnote\relax\footnotetext{ * : equal contribution\\
\indent Project page: \url{thomas-tanay.github.io/grvs}}

\begin{abstract}
Synthesizing novel views from monocular videos of dynamic scenes remains a challenging problem. Scene-specific methods that optimize 4D representations with explicit motion priors often break down in highly dynamic regions where multi-view information is hard to exploit. Diffusion-based approaches that integrate camera control into large pre-trained models can produce visually plausible videos but frequently suffer from geometric inconsistencies across both static and dynamic areas. Both families of methods also require substantial computational resources. Building on the success of generalizable models for static novel view synthesis, we adapt the framework to dynamic inputs and propose a new model with two key components: (1) a recurrent loop that enables unbounded and asynchronous mapping between input and target videos and (2) an efficient use of plane sweeps over dynamic inputs to disentangle camera and scene motion, and achieve fine-grained, six-degrees-of-freedom camera controls. We train and evaluate our model on the UCSD dataset and on Kubric-4D-dyn, a new monocular dynamic dataset featuring longer, higher resolution sequences with more complex scene dynamics than existing alternatives. Our model outperforms four Gaussian Splatting-based scene-specific approaches, as well as two diffusion-based approaches in reconstructing fine-grained geometric details across both static and dynamic regions.
\end{abstract}

\begin{figure}[ht]
    \centering

    \makebox[0.01\columnwidth]{\hspace{-0.2cm} \rotatebox{90}{}}\enspace
    \makebox[\kubricc]{\scriptsize 4DGS~\cite{Wu_2024_CVPR} }\hfill
    \makebox[\kubricc]{\scriptsize Gen3C~\cite{ren2025gen3c}}\hfill
    \makebox[\kubricc]{\scriptsize GRVS (ours)}\hfill
    \makebox[\kubricc]{\scriptsize GT}
    \vspace{0.05cm}
    
    \raisebox{-.5\height}{\makebox[0.01\columnwidth]{\hspace{-0.2cm} \rotatebox{90}{\scriptsize Full}}}\enspace
    \raisebox{-.5\height}{\includegraphics[width=\kubricc,height=\kubricc]{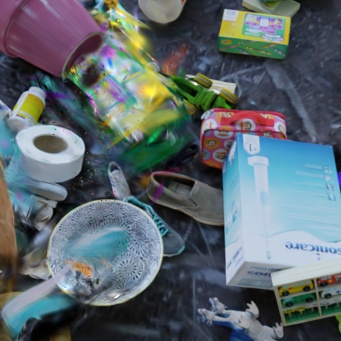}}\hfill
    \raisebox{-.5\height}{\includegraphics[width=\kubricc,height=\kubricc]{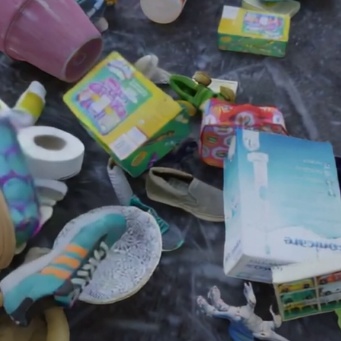}}\hfill
    \raisebox{-.5\height}{\includegraphics[width=\kubricc,height=\kubricc]{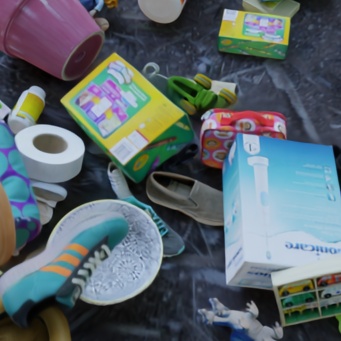}}\hfill
    \raisebox{-.5\height}{\includegraphics[width=\kubricc,height=\kubricc]{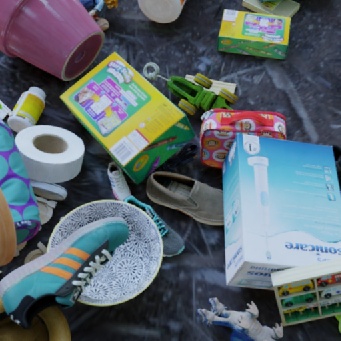}}

    \raisebox{-.5\height}{\makebox[0.01\columnwidth]{\hspace{-0.2cm} \rotatebox{90}{\scriptsize Static}}}\enspace
    \raisebox{-.5\height}{\includegraphics[width=\kubricc,height=\kubricc]{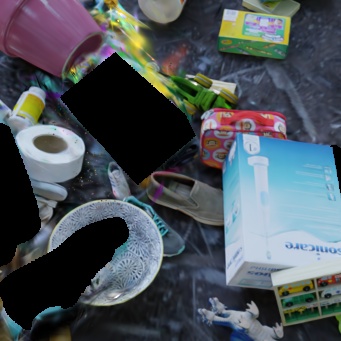}}\hfill
    \raisebox{-.5\height}{\includegraphics[width=\kubricc,height=\kubricc]{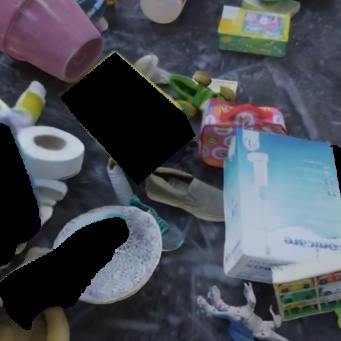}}\hfill
    \raisebox{-.5\height}{\includegraphics[width=\kubricc,height=\kubricc]{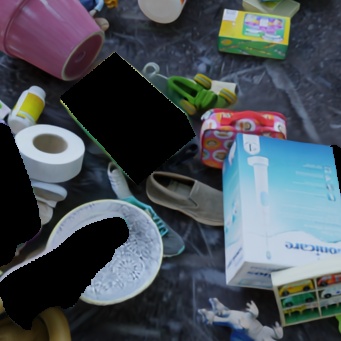}}\hfill
    \raisebox{-.5\height}{\includegraphics[width=\kubricc,height=\kubricc]{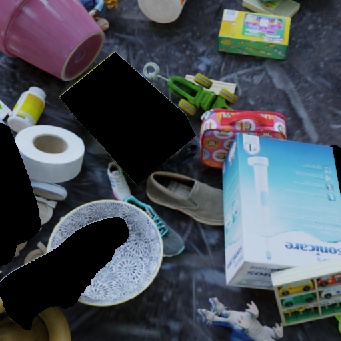}}

    \raisebox{-.5\height}{\makebox[0.01\columnwidth]{\hspace{-0.2cm} \rotatebox{90}{\scriptsize Dynamic}}}\enspace
    \raisebox{-.5\height}{\includegraphics[width=\kubricc,height=\kubricc]{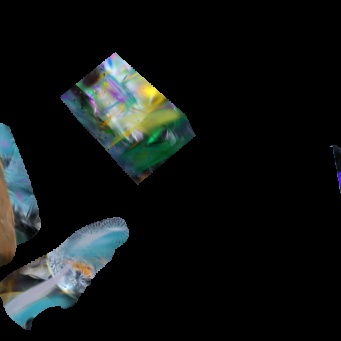}}\hfill
    \raisebox{-.5\height}{\includegraphics[width=\kubricc,height=\kubricc]{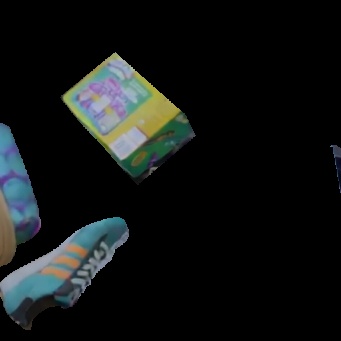}}\hfill
    \raisebox{-.5\height}{\includegraphics[width=\kubricc,height=\kubricc]{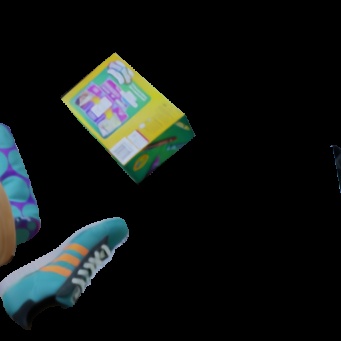}}\hfill
    \raisebox{-.5\height}{\includegraphics[width=\kubricc,height=\kubricc]{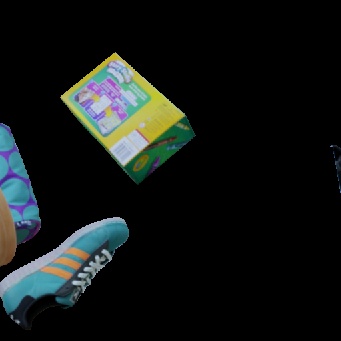}}

    \caption{Scene-specific approaches optimized under motion priors typically reconstruct static regions well but struggle with dynamic elements (\eg 4DGS~\cite{Wu_2024_CVPR}). Diffusion-based approaches conditioned on projected point-clouds typically struggle with fine-grained geometries (\eg Gen3C~\cite{ren2025gen3c}). Our proposed method reconstructs static and dynamic elements with high accuracy.}
    \label{fig:teaser}
\end{figure}

\section{Introduction}
\label{sec:intro}

Novel view synthesis (NVS) methods today often rely on a static scene assumption, where input images are either captured simultaneously by multiple cameras, or sequentially by a single camera observing a scene that remains perfectly still during capture~\cite{mildenhall2020nerf,kerbl3Dgaussians}.
This assumption guarantees 3D-consistent multi-view information but also severely limits the applicability of such methods, since most casually captured videos (\eg handheld mobile recordings) are monocular and contain both static and dynamic elements.

In this paper, we address novel view synthesis from monocular videos of dynamic scenes. This is a particularly challenging problem, because camera and scene motion are often entangled, and multi-view cues are difficult to exploit for dynamic elements.
Current approaches to this problem largely fall into two categories.
The first group of methods optimizes a 4D scene representation---for instance, a static 3D NeRF or Gaussian Splatting model combined with a temporal deformation field---under explicit motion and geometric regularization constraints~\cite{park2021nerfies,li2023dynibar,Wu_2024_CVPR,huang2024sc,yang2024deformable3dgs,lei2024mosca}.
While these methods typically reconstruct static content well, they are computationally expensive (often requiring 20–40 minutes of optimization per scene) and tend to struggle with accurately modeling dynamic elements.
The second line of work leverages large pre-trained video diffusion models, capitalizing on their rich learned video priors.
The key challenge here lies in incorporating camera controls into these models, which is typically achieved either through direct conditioning on camera parameters~\cite{vanhoorick2024gcd,bai2025recammaster} or indirect conditioning via projected point clouds extracted from input images~\cite{mark2025trajectorycrafter,ren2025gen3c}.
Although this latter class of methods currently appears most promising, it still faces several critical limitations.
Diffusion-based approaches that use direct conditioning on camera parameters are typically restricted to a small set of predefined trajectories and can only generate videos synchronized with the input sequences. They provide coarse camera control with only a loose correspondence to the actual scene geometry.
Methods that employ indirect conditioning via projected point clouds, on the other hand, depend heavily on the accuracy of the point-cloud estimation.
For example, the state-of-the-art Gen3C~\cite{ren2025gen3c} relies on ViPE~\cite{huang2025vipe} for point-cloud prediction, which uses monocular depth priors and often produces geometrically inconsistent reconstructions in complex scenes.
Once estimated, such inconsistencies cannot be corrected by the diffusion model (see Figure~\ref{fig:teaser}).
Re-projecting a point cloud is also inherently lossy, leading to degraded fidelity even when source views are available.
Finally, employing a large diffusion model may be unnecessary for monocular dynamic NVS, since most of the visual content in the target views is visible in the input views and the generative demand is often limited to small disoccluded regions.

In this paper, we approach NVS from monocular videos of dynamic scenes as a natural extension of static NVS, and build on recent feedforward models that predict target views directly without relying on explicit scene representations~\cite{tanay24global,jin2024lvsm}.
We begin by re-framing the task as an asynchronous mapping between input and target videos, enabling free-viewpoint rendering in 4D, and introduce a recurrent framework to address this mapping problem. 
The recurrence is beneficial at three complementary levels:
it improves efficiency by continuously reusing previously generated renderings;
it allows the model to leverage both \emph{global} and \emph{local} temporal context by controlling the sampling rate of the inputs; and it ensures that the generated videos remain temporally consistent across successive target views.
To enable novel-view prediction at arbitrary target time steps, we use a training scheme in which each target step is associated with a separately sampled set of input images, and the supervision pairs the target frame with the centrally sampled input frame.
Finally, we extend the use of plane sweep volumes for precise camera controls to dynamic inputs, and argue that this offers an effective mechanism for disentangling camera motion from scene motion, thereby allowing the model to concentrate its capacity on predicting dynamic elements. 

We demonstrate the effectiveness of our approach on the UCSD dataset~\cite{lin2021deep}, as well as on a novel synthetic dataset that we generated using Kubric~\cite{greff2021kubric}. This new dataset consists in longer sequences than previously used~\cite{seitzer2024dyst,vanhoorick2024gcd}, with scenes of higher complexity and at higher resolution ($512\times512$). Our generalizable model is trained end-to-end on RGB images and camera parameters only.
It outperforms state-of-the-art scene-specific gaussian-splatting based approaches, as well as two diffusion based approaches, both quantitatively and qualitatively.

\begin{figure*}[t]
    \centering
    \includegraphics[width=0.9\textwidth]{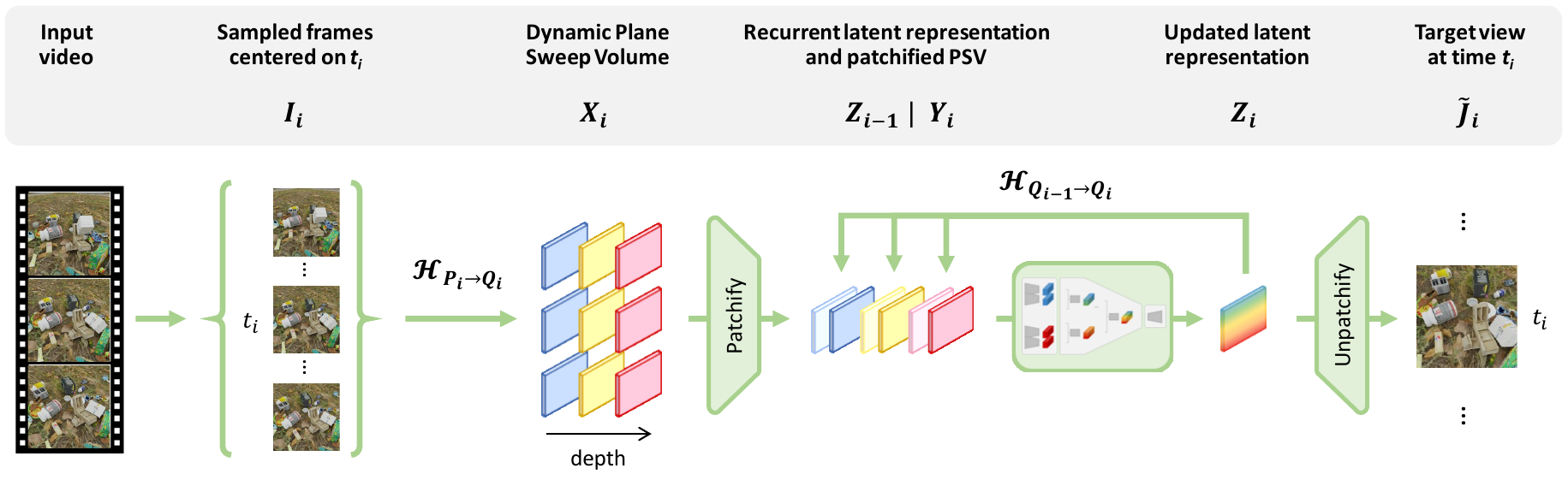}
    \caption{\textbf{GRVS.} For a target view $\mathbf{J}_{i}$ at time $t_i$ and with camera parameters $\mathbf{Q}_{i}\,$, our Generalizable Recurrent View Synthesizer consists of 5 stages. 1) The selection of $V$ input views $\mathbf{I}_{i}$ uniformly sampled around the time $t_i\,$, with corresponding camera parameters $\mathbf{P}_{i}$. 2) The projection of $\mathbf{I}_{i}$ into a dynamic plane sweep volume $\mathbf{X}_{i}$ using the homographies $\mathcal{H}_{\mathbf{P}_{i} \to \mathbf{Q}_{i}}$. 3) The patchification and reshaping of $\mathbf{X}_{i}$ into a downsampled tensor $\mathbf{Y}_{i}$. 4) The latent rendering of $\mathbf{Y}_{i}$ into a hidden state $\mathbf{Z}_{i}$ using the recurrent hidden state $\mathbf{Z}_{i-1}$ projected using the homographies $\mathcal{H}_{\mathbf{Q}_{i-1} \to \mathbf{Q}_{i}}$. 5) The decoding of $\mathbf{Z}_{i}$ into the predicted output $\mathbf{\tilde{J}}_{i}$.}
    \label{fig:GRVS}
\end{figure*}

\section{Related work}
\label{sec:related work}

\paragraph{NVS on static scenes}
Two influential frameworks to perform novel view synthesis from multi-view inputs on static scenes are Neural Radiance Fields (NeRFs)~\cite{mildenhall2020nerf} and 3D Gaussian Splatting (3DGS)~\cite{kerbl3Dgaussians}. These approaches are typically used with a large number of input views in a dense coverage of the scene (\eg 100s or 1000s of views), but they tend to produce degenerate outputs when the number of input views becomes sparse (\eg 10s or less).

A large body of work has explored ways to leverage scene priors to deal with the ill-posedness of sparse NVS. Some methods define explicit priors and  optimize them in a scene-specific manner, for instance by enforcing depth smoothness~\cite{niemeyer2022,wang2023,paliwal2024,xu2024b}, multi-view consistency~\cite{truong2023,kwak2023,paliwal2024,seo2023}, or sparsity~\cite{kim2022,yang2023,seo2023,zhu2024vanilla,zhang2024} constraints. Other methods learn implicit priors over many scenes in a generalizable way. These approaches mainly differ from each other in the type of 3D representations they infer: some methods predict multiplane images~\cite{zhou2018stereo,mildenhall2019local,flynn2019deepview}, some methods predict radiance fields~\cite{chibane2021stereo,yu2021,wang2021ibrnet,mvsnerf21,johari2022geonerf,xu2024}, some methods predict 3D Gaussians~\cite{charatan24pixelsplat,chen2024mvsplat,wewer24latentsplat,chen2024mvsplat360,ye2024noposplat,tang2024lgm,gslrm2024,ziwen2024llrm} and some methods directly predict rendered pixel colors~\cite{suhail2022generalizable,t2023is,cong2023,tanay24global,jin2024lvsm,jiang2025rayzer}. 
These methods also differ from each other in the way multi-view consistency and camera controls are enforced: through epipolar constraints~\cite{chibane2021stereo,yu2021,wang2021ibrnet,t2023is,cong2023,charatan24pixelsplat,wewer24latentsplat}, cost volumes~\cite{mvsnerf21,johari2022geonerf,chen2024mvsplat,chen2024mvsplat360}, plane sweep volumes~\cite{zhou2018stereo,mildenhall2019local,flynn2019deepview,xu2024,tanay24global}, or Pl\"{u}cker embeddings~\cite{suhail2022generalizable,tang2024lgm,gslrm2024,ziwen2024llrm,jin2024lvsm,jiang2025rayzer}. 

The extreme sparsity case of NVS from a single input view has also attracted significant interest recently~\cite{single_view_mpi,szymanowicz24splatter,hong2024lrm,liang2024wonderland,yu2024viewcrafter}. In this case, there is no multi-view information to exploit anymore, and the model becomes fully reliant on depth and scene priors. Impressive results are starting to appear but there remains fundamental limitations in what can be achieved in terms of accurate reconstruction of relative distances and disoccluded content. 

\paragraph{NVS on dynamic scenes}

Many scene-specific approaches have tackled NVS of dynamic scenes from multiview videos. They typically represent the scene with a canonical representation, consisting of a radiance field \cite{wang2022fourier,fridovich2023k,fang2022fast,cao2023hexplane} or using Gaussian Splatting \cite{swings24,duan20244d,yang2024realtime,luiten2023dynamic,lin2024gaussian,li2024spacetime,kratimenos2024dynmf}, and incorporate some motion model deforming the canonical scene to the target timestep. 

Other methods consider monocular videos instead for greater practicability \cite{yoon2020novel,xian2021space,tretschk2021non,pumarola2021d,li2021neural,park2021nerfies,Gao-ICCV-DynNeRF,park2021hypernerf,wu2022d,gao2022monocular,
liu2023robust,Wu_2024_CVPR,huang2024sc,yang2024deformable3dgs,lee2024fast}, which dramatically increases the complexity of the task. They resort to hand-crafted priors to further constrain the problem. Among them, 4DGS~\cite{Wu_2024_CVPR} uses a total variation loss in order to regularize a voxel feature grid, which is decoded into an
offset of the Gaussians' attributes modeling their motion. SC-GS~\cite{huang2024sc} models the motion using a small set of control points associated with a rigid body motion basis. 
D-3DGS~\cite{yang2024deformable3dgs} regularizes the motion by representing the deformation using an MLP, making use of its inherent smoothness. 
Building on the need for structured motion, MoSca~\cite{lei2024mosca} extracts a sparse, graph-based motion scaffold via 2D foundation models to smoothly constrain and propagate Gaussian deformations across time-frames.
4DGS, SC-GS, D-3DGS and MoSca are four scene-specific baselines we compare our approach to in Section~\ref{sec:Results}. The results show that they reconstruct static regions well, but struggle with dynamic elements. 
Recent works have also explored the use of additional priors that are obtained from large pre-trained models to improve the results of scene-specific optimizations. This includes image and video diffusion \cite{wu2024cat4d,wang2024diffusion,nazarczuk2024vidar}, or depth map, segmentation mask, and long-term 2D tracks \cite{li2023dynibar,lei2024mosca,som2024,jeong2024rodygs}. 
However, these approaches require to perform a lengthy optimization for every single scene and hand-crafted priors often fail to capture the complexity and diversity of real scenes. 

Generalizable approaches to monocular dynamic NVS are less explored~\cite{vanhoorick2024gcd,seitzer2024dyst,ren2024l4gm,ren2025gen3c,bai2025recammaster,mark2025trajectorycrafter}. 
A recent trend consists in integrating camera controls into pre-trained diffusion models, either through direct conditioning, by injecting camera pose information inside the diffusion model~\cite{vanhoorick2024gcd,bai2025recammaster}, or through explicit point cloud re-projections, by fine-tuning the diffusion model on such inputs~\cite{mark2025trajectorycrafter,ren2025gen3c}. 
GCD~\cite{vanhoorick2024gcd} and Gen3C~\cite{ren2025gen3c}, in particular, are the two diffusion-based baselines we compare our approach to in Section~\ref{sec:Results}. The results show that they both struggle with fine-grained geometries, irrespectively of the type of camera control used. 

Our approach also differs from the aforementioned works in its use of a recurrent architecture.
We draw inspiration for this recurrent architecture from iterative approaches to optical flow estimation and SLAM~\cite{teed2020RAFT,wang2024sea,teed2021droid,li2024_MegaSaM}, as well as  recurrent approaches to video denoising and super-resolution~\cite{sajjadi2018frame,godard18deep,chan2021basicvsr,chan2022basicvsrpp,tanay2022diagnosing}.

\input{figures/DPVS}

\section{Method}

We present a new method for monocular dynamic NVS that builds on existing generalizable methods for static NVS.

\subsection{NVS on static scenes}

Consider a set of $V$ input views of a \emph{static scene}, consisting of color images and camera parameters. The images are of height $H$ and width $W$, with red-green-blue color channels, and can be stacked into a 4D tensor $\mathbf{I} \in \mathbb{R}^{V \times 3 \times H \times W}$. The camera parameters consist of intrinsics and extrinsics and can be stacked into a pair of tensors $\mathbf{P} \in \mathbb{R}^{V \times 3 \times 3} \times \mathbb{R}^{V \times 3 \times 4}$. Now consider a distinct target view with ground-truth image $\mathbf{J} \in \mathbb{R}^{3 \times H \times W}$, and camera parameters  $\mathbf{Q} \in \mathbb{R}^{3 \times 3} \times \mathbb{R}^{3 \times 4}$. We are interested in \emph{novel view synthesis}, which consists in predicting an estimate $\mathbf{\tilde{J}}$ of the target image $\mathbf{J}$, given the input images $\mathbf{I}$, and the camera parameters $\mathbf{P}$ and $\mathbf{Q}$. 

Following ConvGLR~\cite{tanay24global} and LVSM~\cite{jin2024lvsm}, we learn a generalizable model $\mathcal{M}$ that solves this task directly, without predicting any intermediary 3D representation:
\begin{equation}
    \mathbf{\tilde{J}} = \mathcal{M}(\mathbf{I}, \mathbf{P}, \mathbf{Q}) \,.
\label{static_eq}
\end{equation}
We specifically build on the approach of \mbox{ConvGLR}~\cite{tanay24global}.
First, the camera pose information is encoded at the image level by projecting the input images into a plane sweep volume (PSV) at the target camera position, using homographies computed from the input and target camera parameters:
$\mathbf{X} = \mathcal{H}_{\mathbf{P} \to \mathbf{Q}}(\mathbf{I})$.
Then, the target image is predicted from the PSV: $\mathbf{\tilde{J}} = \mathcal{M}(\mathbf{X})$, where the model $\mathcal{M}$ is an efficient 3D Unet architecture that performs the rendering operation globally in a low-resolution latent space.

\subsection{NVS on dynamic scenes}

Now, consider a monocular video of a \emph{dynamic scene} consisting of $T$ input views captured in successive time steps for $t \in [1,T]$. We are interested in predicting a novel video of the same scene, consisting of $N$ target views for $i \in [1,N]$. 
In order to allow free viewpoint rendering in 4D, we assume here that $t$ and $i$ are distinct iterators, and the relationship between the two can be set arbitrarily by the user by defining a mapping $i \to t_i$, subject only to the temporal constraint: $t_i \in \{t_{i-1}-1\,,\,t_{i-1}\,,\,t_{i-1}+1\}$. 
This is in contrast with the approach adopted in GCD~\cite{vanhoorick2024gcd} for instance, where the mapping between input and target views is synchronous by construction, with $t_i = i$ for all $i$. 
An asynchronous mapping that is of specific interest is the case of bullet-time videos, where $t_i$ is constant for all $i$ and the dynamic scene appears frozen in time in the output video.

We approach this new problem by making a number of key modifications to the static NVS framework.
First, we warp the generalizable model into a recurrent loop, 
predicting target video frames $\mathbf{\tilde{J}}_{i}$ successively while propagating a recurrent hidden state $\mathbf{Z}_{i}$ between frames:
\begin{equation}
\mathbf{\tilde{J}}_{i} \,,\, \mathbf{Z}_{i} = \mathcal{M}(\mathbf{I}_{i}\,,\,\mathbf{P}_{i}\,,\,\mathbf{Q}_{i}\,,\,\mathbf{Z}_{i-1}) 
\label{dynamic_eq}
\end{equation}
for $i \in [1,N]$ with $\mathbf{Z}_0 = 0$. 
We also add patchification and unpatchification modules around the model to gain in computational efficiency and allow its unrolling through time during training. 
Finally we train the model to predict novel views at chosen target time steps by sampling a different set of input images $\mathbf{I}_{i}$ for each target time step $t_i$, and by pairing the target image $\mathbf{J}_{i}$ with the middle frame in $\mathbf{I}_{i}$. 
Our Generalizable Recurrent View Synthesizer (GRVS) consists of the following 5 main stages (see also Figure~\ref{fig:GRVS}). 

\paragraph{Input selection}
For an output frame $i$, we start by fixing a target time step $t_i$ and sample an odd subset of $V$ input images $\mathbf{I}_{i} \in \mathbb{R}^{V \times 3 \times H \times W}$ from the input video.
The images in $\mathbf{I}_{i}$ are uniformly sampled around $t_i$ with a dilation factor $d$, \ie they consist in the video frames with indices in $\{\,\cdots\,,\, t_{i}-2d\,,\,t_{i}-d\,,\,t_i\,,\,t_i+d,\,t_i+2d\,,\,\cdots\}$.
A large value of $d$ provides the model with \emph{global context} around $t_i$, improving the reconstruction of static regions, while a small value of $d$ provides the model with \emph{local context} around $t_i$, improving the reconstruction of dynamic regions. 
In practice, it is possible to operate the model in \emph{iterative} mode, where multiple passes for each target view are performed with a progressively shrinking dilation factor. We show in Section~\ref{sec:ablations} that this allows the model to achieve optimal performance by accumulating both global and local context.

\paragraph{Dynamic Plane Sweep Volume}
We then project the input images into a PSV $\mathbf{X}_{i}$, using homographies computed from the input and target camera parameters: $\mathbf{X}_{i} = \mathcal{H}_{\mathbf{P}_{i} \to \mathbf{Q}_{i}}\,(\mathbf{I}_{i})$. 
Like in the case of static NVS, this PSV serves as an encoding of the relative camera positions between the input views and the target view.
It allows direct camera control and replaces alternatives like conditioning a diffusion model~\cite{vanhoorick2024gcd,bai2025recammaster}, reprojecting a point cloud~\cite{mark2025trajectorycrafter,ren2025gen3c}, or using Pl\"{u}cker embeddings~\cite{jin2024lvsm,jiang2025rayzer}.
Using a PSV encoding is particularly suitable for monocular dynamic NVS, because it helps disentangle camera and scene motion and allows the model to focus on dynamic elements (see Figure~\ref{fig:DPSV}). The dynamic plane sweep volume is a 5D tensor $\mathbf{X}_{i} \in \mathbb{R}^{D \times V \times 3 \times H \times W}$ where $D$ is a hyper-parameter controlling the number of depth planes used. 
 
\paragraph{Patchify} 
We then view the PSV as a 4D tensor of shape $D\!\times\!3V\!\times\!H\!\times\!W$ and feed it to a convolutional layer with kernel size $F$ and stride $F$, effectively patchifying it into a tensor $\mathbf{Y}_{i} \in \mathbb{R}^{D \times C \times \frac{H}{F} \times \frac{W}{F}}$, where $F$ is a hyper-parameter controlling the down-sampling factor, and $C$ is a hyper-parameter controlling the number of channels.
This patchification has a large positive impact on the computational cost of the model (setting $F=2$ roughly divides the FLOP count by 4) but a reasonable negative impact on performance (see Table~\ref{table:Ablations}). It allows us to process high-resolution images or train the model with multiple recurrent iterations unrolled in one batch.

\paragraph{Recurrent Latent Rendering} 
We then use the efficient 3D Unet introduced in ConvGLR~\cite{tanay24global} to produce a latent representation $\mathbf{Z}_{i} \in \mathbb{R}^{1 \times C \times \frac{H}{F} \times \frac{W}{F}}$.
Contrary to~\cite{tanay24global}, however, the 3D Unet now takes as input a concatenation of the current patchified PSV $\mathbf{Y}_{i}$ and the recurrent latent representation $\mathbf{Z}_{i-1}$ re-projected into the current target camera viewpoint using homographies $\mathcal{H}_{\mathbf{Q}_{i-1}\to\mathbf{Q}_{i}}$ computed from the previous and current target camera parameters.
This operation effectively renders the scene at time $t_i$ from the target viewpoint $\mathbf{Q}_{i}$ while efficiently reusing prior renderings through the recurrence. 

\paragraph{Unpatchify} 
Finally, we feed the latent scene representation to a convolutional layer with kernel size $1$ and number of channels $3\times F \times F$, and unpatchify it into the estimated target image $\mathbf{\tilde{J}}_{i} \in \mathbb{R}^{1 \times 3 \times H \times W}$ at time $t_i$, corresponding to the time step of the middle input frame.

\input{figures/UCSD.tex}

\input{figures/Kubric.tex}

\subsection{Datasets}

Training a generalizable model for monocular dynamic view synthesis is challenging due to the large amount of training data required, and its difficult capture.
It requires synchronized multi-view recordings of dynamic scenes using at least two cameras providing input and target videos. 
Accurate synchronization is crucial, or the training signal becomes corrupted by misalignments.
The input camera should be dynamic, and the relative pose between the input and target cameras should vary randomly over the dataset, or the model would fail to generalize to novel relative poses (\ie a fixed rig with 2 cameras would not suffice). Finally, camera poses should be accurately estimated in the presence of large amounts of dynamic distractors. 

Two distinct strategies have been used in prior work to deal with this challenge. One solution consists in using a pre-calibrated static rig with several cameras, and simulate monocular dynamic inputs by alternating samples from different cameras. This protocol was introduced in NSFF~\cite{li2021neural} on the 8 scenes of the Nvidia Dynamic Scenes Dataset~\cite{yoon2020novel} and used again in DynIBaR~\cite{li2023dynibar} on the 10 test scenes of the UCSD dataset~\cite{lin2021deep}, both times in the context of scene-specific approaches. However, the realism and diversity of camera motions that can be synthesized in this way is very limited~\cite{gao2022monocular} and capturing a large training dataset remains challenging. A second solution adopted in more recent works~\cite{seitzer2024dyst,vanhoorick2024gcd} and facilitated by the development of the Kubric dataset generator~\cite{greff2021kubric} is to rely on fully synthetic data generated with a graphics engine. 

In this work we use both approaches. We train our generalizable model on the 86 scenes of the UCSD dataset, and replicate the DynIBaR protocol for evaluation on the 10 test scenes. We also train and evaluate our model on new synthetic data generated with Kubric~\cite{greff2021kubric}.
There already exist two Kubric-based dataset variants for dynamic NVS, but they do not fit our requirements.
The dataset used in DyST~\cite{seitzer2024dyst} has a very low resolution (128$\times$128) and a single object per scene. The dataset used in GCD~\cite{vanhoorick2024gcd} (3000 scenes, 60 frames at 24 FPS) has a higher resolution (576$\times$384) and more scene diversity (7 to 22 objects) but only includes static cameras. 
This is an issue because most 4D reconstruction methods such as D-3DGS~\cite{yang2024deformable3dgs}, SC-GS~\cite{huang2024sc}, 4DGS~\cite{Wu_2024_CVPR}, MoSca~\cite{lei2024mosca} as well as GRVS rely on multi-view cues to estimative the scene geometry.

We therefore generated a new Kubric variant, denoted \emph{Kubric-4D-dyn},  with $5000$ training scenes and $100$ test scenes of 81 frames each, at 512$\times$512 resolution with a frame rate of 24 FPS. Each training scene has $3$ cameras: one dynamic input camera moving over a virtual semi-sphere (camera positions interpolated in spherical coordinates between two random points), and two static target cameras. We populated each scene with 40 static objects and 5 dynamic objects that were given a moderate initial velocity. We set the friction to the minimum so that dynamic objects continue to move during the entire duration of each sequence. For evaluation, we generated 100 scenes under the same conditions using a distinct set of objects and backgrounds to prevent overfitting.

\subsection{Training}

Our model depends on 5 hyper-parameters: the number of channels $C$, the number of depth planes $D$, the downsampling factor $F$, the number of input views $V$ and the dilation factor $d$. We also operate our model in \emph{iterative} mode by doing multiple passes for each target frame with a progressively shrinking dilation factor. For our main model, we set $C=256$, $D=32$, $F=2$, $V=15$ and we perform three successive iterations with $d \in \{5,3,1\}$. 

During training, the successive iterations are unrolled and the recurrent hidden state $\mathbf{Z}_{i}$ is detached from the computation graph between iterations to avoid saturating the GPU memory. The model is trained end-to-end on RGB images only (patches of 320$\times$320 pixels), using a standard VGG loss switched to an L1 loss in the last 10\% of the training to avoid gridding artifacts. We use the Adam optimizer for 100k steps, with a batch size of 4. Training takes around 4 days using 4 GPUs with 32GB of VRAM each.

\section{Results}
\label{sec:Results}

We present quantitative and qualitative results on two datasets, as well as an ablation of the model's recurrence.

\subsection{UCSD}

The UCSD dataset~\cite{lin2021deep} consists of 96 scenes (86 for training and 10 for testing) captured with a static rig of 10 forward facing cameras disposed in a three-row grid (3-4-3), recoding 1-2 minutes of video per scene. Images are downscaled to 640$\times$360 resolution, and monocular sequences are simulated by alternating frames between cameras. Evaluation is performed on all the frames that are not in the input sequence for a number of randomly chosen time steps per scene.

We compare our model to three recent methods based on Gaussian Splatting (GS) for monocular dynamic novel view synthesis: D-3DGS~\cite{yang2024deformable3dgs}, SC-GS~\cite{huang2024sc} and 4DGS~\cite{Wu_2024_CVPR}. All three methods require scene-specific training, which takes around 25 to 50 minutes per scene, while our model is generalizable and can predict novel views on the test scenes directly. We report the performance of the different methods in terms of PSNR, SSIM and LPIPS computed on full images as well as on the dynamic parts only in Table~\ref{table:UCSD} and show some predicted outputs for qualitative evaluation in Figure~\ref{fig:qualitative-results-UCSD}. For reference, we also reproduce in Table~\ref{table:UCSD} the metrics as reported in~\cite{li2023dynibar} for Nerfies~\cite{park2021nerfies}, HyperNeRF~\cite{park2021hypernerf}, DVS~\cite{Gao-ICCV-DynNeRF}, NSFF~\cite{li2021neural} and DynIBaR~\cite{li2023dynibar}.

4DGS, D-3DGS and SC-GS struggle on this task, especially on dynamic regions but also to a lesser extent on static regions, because of the sparse nature of the synthetic sequences used, where only 10 distinct viewpoints are available throughout the sequences (since there are only 10 static cameras). Interestingly, 4DGS performs worse than D-3DGS and SC-GS over the full images, but significantly better over the dynamic regions. Our approach is on-par with DynIBaR in terms of performance both over static and dynamic regions, while being orders of magnitude faster to run. Optimizing a full system on a 10 second video takes around two days using 8 Nvidia A100s for DynIBaR, while our model does not require any scene-specific tuning and renders multiple novel views per seconds. DynIBaR also uses a lot of explicit priors in the form of disparity maps, optical flows, static and dynamic masks, virtual source views, whereas our method only uses learned implicit priors and predicts novel views directly from the input RGB images and their camera parameters. 

\begin{table}[t]
\centering
\begin{adjustbox}{width=\columnwidth}
\begin{tblr}{colspec={Q[2.6cm]|ccc|ccc}}
\toprule[1.5pt]
\SetCell[r=2]{l}{Method} & \SetCell[c=3]{c}{Full image} &  &  & \SetCell[c=3]{c}{Dynamic only} &  &  \\
\cmidrule[lr]{2-4} \cmidrule[lr]{5-7}
 & PSNR$\uparrow$ & SSIM$\uparrow$ & LPIPS$\downarrow$ & PSNR$\uparrow$ & SSIM$\uparrow$ & LPIPS$\downarrow$ \\
\hline\hline
Nerfies*~\cite{park2021nerfies} & 24.32 & 0.823 & 0.096 & 18.45 & 0.595 & 0.234  \\
HyperNeRF*~\cite{park2021hypernerf} & 25.10 & 0.859 & 0.095 & 19.26 & 0.618 & 0.212 \\
DVS*~\cite{Gao-ICCV-DynNeRF} & 30.64 & 0.943 & 0.075 & 26.57 & 0.866 & 0.096 \\
NSFF*~\cite{li2021neural} & 31.75 & 0.952 & 0.034 & 25.83 & 0.851 & 0.115 \\
DynIBaR*~\cite{li2023dynibar} & \SetCell{orange!25} 36.47 & \SetCell{red!25} 0.983 &  \SetCell{orange!25} 0.014 & \SetCell{orange!25} 28.01 & \SetCell{red!25} 0.909 & \SetCell{red!25} 0.042 \\
\hline
D-3DGS~\cite{yang2024deformable3dgs} & 29.98 & 0.906 & 0.080 & 18.17 & 0.409 & 0.364 \\
SC-GS~\cite{huang2024sc} & 28.97 & 0.886 & 0.099 & 18.34 & 0.421 & 0.376 \\
4DGS~\cite{Wu_2024_CVPR} & 25.37 & 0.770 & 0.108 & 22.24 & 0.584 & 0.193  \\
GRVS (ours) & \SetCell{red!25} 36.81 & \SetCell{orange!25} 0.972 & \SetCell{red!25} 0.012 & \SetCell{red!25} 29.12 & \SetCell{orange!25} 0.870 & \SetCell{orange!25} 0.058 \\
\bottomrule[1.5pt]
\end{tblr}
\end{adjustbox}
\caption{\textbf{Quantitative evaluation on UCSD.} 
The metrics are computed on full images (left) and on dynamic regions only (right). Dynamic regions are determined using the dynamic masks provided in~\cite{lin2021deep}. The performances for the starred methods are reproduced from~\cite{li2023dynibar}. For each metric, 1st, and 2nd best-performing methods are highlighted in red and orange respectively.}
\label{table:UCSD}
\end{table}

\begin{table}[t]
\centering
\begin{adjustbox}{width=\columnwidth}
\begin{tblr}{colspec={Q[2.6cm]|ccc|ccc}}
\toprule[1.5pt]
\SetCell[r=2]{l}{Method} & \SetCell[c=3]{c}{Full image} &  &  & \SetCell[c=3]{c}{Dynamic only} &  &  \\
\cmidrule[lr]{2-4} \cmidrule[lr]{5-7}
 & PSNR$\uparrow$ & SSIM$\uparrow$ & LPIPS$\downarrow$ & PSNR$\uparrow$ & SSIM$\uparrow$ & LPIPS$\downarrow$ \\
\hline\hline
\SetCell{gray!10} & \SetCell[c=6]{gray!10}{target 1} & & & & &  \\
\hline
D-3DGS~\cite{yang2024deformable3dgs} & 23.71 & 0.861 & 0.173 & 15.69 & 0.349 & 0.449 \\
SC-GS~\cite{huang2024sc} & \SetCell{orange!25} 24.24 & \SetCell{orange!25} 0.862 & \SetCell{orange!25} 0.164 & 16.59 & 0.401 & 0.396 \\
4DGS~\cite{Wu_2024_CVPR} & 23.70 & 0.852 & 0.189 &  15.69 & 0.344 & 0.449 \\
MoSca~\cite{lei2024mosca} & 22.56 & 0.777 & 0.217 & 17.76 & 0.514 & 0.427 \\
GCD~\cite{vanhoorick2024gcd} & 18.00 & 0.448 & 0.436 &  17.13 & 0.382 & 0.446 \\
Gen3C~\cite{ren2025gen3c} & 19.31  & 0.591  & 0.343  & \SetCell{orange!25} 18.28  & \SetCell{orange!25} 0.530  & \SetCell{orange!25} 0.389  \\
GRVS (ours) & \SetCell{red!25} 28.42 & \SetCell{red!25} 0.913 & \SetCell{red!25} 0.113 & \SetCell{red!25} 22.27 & \SetCell{red!25} 0.724 & \SetCell{red!25} 0.177 \\
\hline
\SetCell{gray!10} & \SetCell[c=6]{gray!10}{target 2} & & & & &  \\
\hline
D-3DGS~\cite{yang2024deformable3dgs} & 21.40 & \SetCell{orange!25} 0.796 & 0.229 & 13.89 & 0.259 & 0.528 \\
SC-GS~\cite{huang2024sc} & \SetCell{orange!25} 21.41 & 0.781 & \SetCell{orange!25} 0.230 & 14.55 & 0.290 & 0.499 \\
4DGS~\cite{Wu_2024_CVPR} & 21.35 & 0.787 & 0.246 & 13.91 & 0.251 & 0.531  \\
MoSca~\cite{lei2024mosca} & 20.34 & 0.722 & 0.282 & 15.44 & \SetCell{orange!25} 0.434 & 0.523 \\
GCD~\cite{vanhoorick2024gcd} & 16.48 & 0.400 & 0.505 &  15.78 & 0.325 & 0.519 \\
Gen3C~\cite{ren2025gen3c} & 16.89 & 0.492  & 0.447  & \SetCell{orange!25} 15.91  & 0.426  & \SetCell{orange!25} 0.491  \\
GRVS (ours) & \SetCell{red!25} 25.53 & \SetCell{red!25} 0.866 & \SetCell{red!25} 0.161 & \SetCell{red!25} 19.10 & \SetCell{red!25} 0.569 & \SetCell{red!25} 0.281 \\
\hline
\SetCell{gray!10} & \SetCell[c=6]{gray!10}{target 3} & & & & &  \\
\hline
D-3DGS~\cite{yang2024deformable3dgs} & 19.00 & \SetCell{orange!25} 0.708 & \SetCell{orange!25} 0.286 & 13.28 & 0.232 & 0.561 \\
SC-GS~\cite{huang2024sc} & 18.76 & 0.680 & 0.299 & 13.71 & 0.252 & 0.553 \\
4DGS~\cite{Wu_2024_CVPR} & \SetCell{orange!25} 19.30 & \SetCell{orange!25} 0.708 & 0.300 & 13.26 & 0.227 & 0.569  \\
MoSca~\cite{lei2024mosca} & 18.65 & 0.663 & 0.352 & 14.60 & \SetCell{orange!25} 0.393 & 0.571 \\
GCD~\cite{vanhoorick2024gcd} & 15.23 & 0.374 & 0.562 & \SetCell{orange!25} 14.72 & 0.293 & 0.580 \\
Gen3C~\cite{ren2025gen3c} & 15.59  & 0.448  & 0.512  & 14.70  & 0.384  & \SetCell{orange!25} 0.553  \\
GRVS (ours) & \SetCell{red!25} 22.78 & \SetCell{red!25} 0.798 & \SetCell{red!25} 0.220 & \SetCell{red!25} 17.37 & \SetCell{red!25} 0.480 & \SetCell{red!25} 0.370 \\
\bottomrule[1.5pt]
\end{tblr}
\end{adjustbox}
\caption{\textbf{Quantitative evaluation on Kubric-4D-dyn.} The metrics are computed on full images (left) and on dynamic regions only (right), for three targets at increasing distances from the input trajectory. Full images are center-cropped to match the rectangular dimensions of GCD predictions (341$\times$512). Dynamic regions are determined by segmenting out dynamic objects. For each metric, 1st, and 2nd best-performing methods are highlighted in red and orange respectively.}
\label{table:Kubric}
\end{table}

\subsection{Kubric-4D-dyn}

For this set of experiments, we compare our model to the same three GS-based methods as for UCSD as well as MoSca~\cite{lei2024mosca}, and two recent diffusion-based approaches: GCD~\cite{vanhoorick2024gcd} and Gen3C~\cite{ren2025gen3c}.
For GCD, we use the publicly available model weights trained on Kubric-4D in gradual mode with a maximum azimuth of 90 degrees (best performing model). 
For Gen3C, we follow the recommended protocol for video to video generation and extract camera poses and depth maps using ViPE~\cite{huang2025vipe}. 
When computing full image metrics, we center-crop the generated outputs of all the models to match the rectangular dimensions of GCD outputs (\ie 341$\times$512). Again, D-3DGS~\cite{yang2024deformable3dgs}, SC-GS~\cite{huang2024sc}, 4DGS~\cite{Wu_2024_CVPR} as well as MoSca~\cite{lei2024mosca} require scene-specific training, which takes around 20 to 40 minutes per scene, 
while GCD, Gen3C and our model can predict novel views on the test scenes directly. All the models are evaluated in three scenarios, with target views placed at increasing distances from the input trajectory. We report the performances of the different methods in terms of PSNR, SSIM and LPIPS on full images as well as on the dynamic parts only in Table~\ref{table:Kubric} and show some predicted outputs for qualitative evaluation in Figure~\ref{fig:qualitative-results-Kubric}.

The four GS-based methods reconstruct static regions well but struggle with dynamic objects, especially when the distance between the input trajectory and the target viewpoint increases.
MoSca~\cite{lei2024mosca} is noticeably better on dynamic objects, but this comes with a slightly reduced performance on full images. 
On the contrary, the two diffusion-based methods predict static and dynamic objects that are plausible, but with inaccurate 3D geometries, image textures, and fine-grained camera positions.
This is true both for GCD, which integrates camera controls into Stable Video Diffusion~\cite{blattmann2023stable} through micro-conditioning, and for Gen3C, which relies on point cloud estimates from ViPE~\cite{huang2025vipe}, which contain significant geometric inaccuracies. 

Our model outperforms the baselines for the three metrics, both on full images and on dynamic regions only. Like the GS-based models, it predicts static regions with high accuracy, thanks to effective exploitation of multi-view information obtained from the computation of plane sweep volumes. 
Like the diffusion-based models, it also predicts dynamic regions remarkably well, thanks to its generalizable nature and the fact that it was trained on a large amount of monocular dynamic sequences. 

\begin{figure}[t]
\centering
\includegraphics[width=\columnwidth]{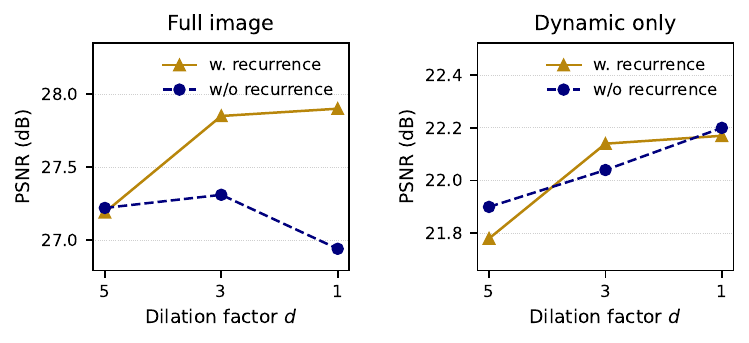}
\caption{\textbf{Recurrence}. PSNR on target 1 over three iterations with decreasing dilation factors $d \in \{5,3,1\}$ for two model variants, trained and evaluated with and without recurrence.}
\label{fig:iterations}
\end{figure}

\subsection{Ablations}
\label{sec:ablations}

In this section, we study the influence of a number of hyper-parameters on the performance of our model. 
In particular, we motivate the use of a recurrent architecture by training two model variants: one with the recurrence enabled, and the other one without. We then evaluate the performance of both models on target 1 over three iterations while varying the dilation factor $d \in \{5,3,1\}$ (see Figure~\ref{fig:iterations}).

As hypothesized earlier, the performance of the model trained and evaluated without recurrence decreases on full images and increases on dynamic regions when $d$ decreases. This is because the model receives more global temporal information when $d$ is large (beneficial for static regions), and more local temporal information when $d$ is small (beneficial for dynamic regions). In contract, the performance of the model trained with recurrence increases both on full images and on dynamic regions when $d$ decreases. This demonstrates that the recurrent hidden state propagates meaningful information temporally: on the third iteration, the model trained with recurrence outperforms the model trained without recurrence by around 1 dB on full images, while keeping the same performance on dynamic regions. The use of the recurrence is an effective way to incorporate both global and local temporal context. 

We then consider the influence of other hyper-parameters in Table~\ref{table:Ablations}. Starting from a model with $C=128$, $D=32$, $F=2$, $V=9$ and three iterations with $d \in \{9,5,1\}$, we vary the number of depth planes $D \in \{16,64\}$, the down-sampling factor $F \in \{1,4\}$, and the number of input views $V \in \{3,15\}$ (while simultaneously adapting the dilation factor to roughly cover similar global and local temporal regions). When increasing the number of depth planes to $D=64$, we keep the backbone 3D Unet the same by grouping the depth planes by 2, as done in~\cite{tanay24global}. We did not observe any performance improvement by doing this on this task, but reducing the number of depth planes to $16$ had a moderate negative impact on performance, while significantly increasing the FPS. 
Increasing the down-sampling factor $F$ has a negative impact on performance, but it significantly increases the FPS and therefore reduces the training time. Using a larger value of $F$ can also be particularly beneficial when working on high-resolution images. Finally, increasing the number of views $V$ has a positive impact on performance, but the model trained with only three views still performs remarkably well given very sparse inputs. 

\begin{table}[t]
\centering
\begin{adjustbox}{width=\columnwidth}
\begin{tblr}{ccccc|ccc|cc}
\toprule[1.5pt]
\SetCell[c=5]{c} Hyper-parameters & & & & & \SetCell[c=3]{c}{Full image} &  &  & \SetCell[r=2]{c}{FPS} & \SetCell[r=2]{c}{params} \\
\cmidrule[lr]{1-5} \cmidrule[lr]{6-8}
$C$ & $D$ & $F$ & $V$ & $d$ & PSNR$\uparrow$ & SSIM$\uparrow$ & LPIPS$\downarrow$ & & \\
\hline\hline
128 & \textbf{16} & 2 & 9 & 9-5-1 & 27.17 & 0.885 & 0.144 & 4.4 & 38M \\
128 & \textbf{32} & 2 & 9 & 9-5-1 & 27.58 & 0.895 & 0.134 & 3.0 & 40M \\
128 & \textbf{64} & 2 & 9 & 9-5-1 & 27.53 & 0.895 & 0.134 & 2.3 & 40M \\
\hline
128 & 32 & \textbf{1} & 9 & 9-5-1 & 28.80 & 0.926 & 0.096 & 1.2 & 40M \\
128 & 32 & \textbf{2} & 9 & 9-5-1 & 27.58 & 0.895 & 0.134 & 3.0 & 40M \\
\textbf{256} & 32 & \textbf{4} & 9 & 9-5-1 & 26.55 & 0.855 & 0.176 & 3.4 & 161M \\
\hline
128 & 32 & 2 & \textbf{3} & \textbf{30-10-3} & 27.41 & 0.889 & 0.144 & 3.7 & 40M \\
128 & 32 & 2 & \textbf{9} & \textbf{9-5-1} & 27.58 & 0.895 & 0.134 & 3.0 & 40M \\
128 & 32 & 2 & \textbf{15} & \textbf{5-3-1} & 27.90 & 0.900 & 0.095 & 2.5 & 40M \\
\bottomrule[1.5pt]
\end{tblr}
\end{adjustbox}
\caption{\textbf{Influence of various hyper-parameters.} Each model is trained on Kubric-4D-dyn and evaluated on target 1. The frame rate (FPS) and the number of parameters are also reported. Performance increases with the number of depth planes $D$ and the number of input views $V$, and decreases with the patch size $F$.}
\label{table:Ablations}
\end{table}

\section{Conclusion}

We introduce GRVS, a new generalizable approach for monocular dynamic NVS that overcomes the limitations of both optimization-based 4D reconstruction pipelines and diffusion-based methods relying on imprecise geometric priors.
Our recurrent architecture enables efficient reuse of intermediate predictions, robust aggregation of temporal context, and free-viewpoint rendering in 4D, while our use of plane sweep volumes provides a lightweight and effective mechanism for separating camera motion from scene motion.
Unlike diffusion-based approaches, GRVS operates without external priors or pre-trained modules and remains substantially more lightweight (40M parameters and 3FPS for GRVS vs 7B parameters and 0.3FPS for Gen3C~\cite{ren2025gen3c}).
Extensive experiments on both the UCSD dataset and our new high-resolution Kubric-4D-dyn dataset show that our method generalizes across scenes and outperforms existing dynamic NVS approaches in both quantitative accuracy and perceptual quality.
Together, these results suggest a promising alternative to current diffusion-centric strategies and point toward efficient, geometry-aware models as a viable path forward for real-world dynamic view synthesis.

{
    \small
    \bibliographystyle{ieeenat_fullname}
    \bibliography{main}
}

\end{document}

%% file: preamble.tex
\usepackage{float}
\restylefloat{table}
\usepackage{adjustbox}
\usepackage{tabularray}
\UseTblrLibrary{booktabs}
\usepackage{booktabs}
\usepackage{diagbox}
\usepackage{bm}
\usepackage{caption}

\usepackage{microtype}



%% file: figures/DPVS.tex
\begin{figure*}[t]
    \centering

    \makebox[\textwidth]{\scriptsize Static scene}\vspace{0.05cm}
    \includegraphics[width=\textwidth]{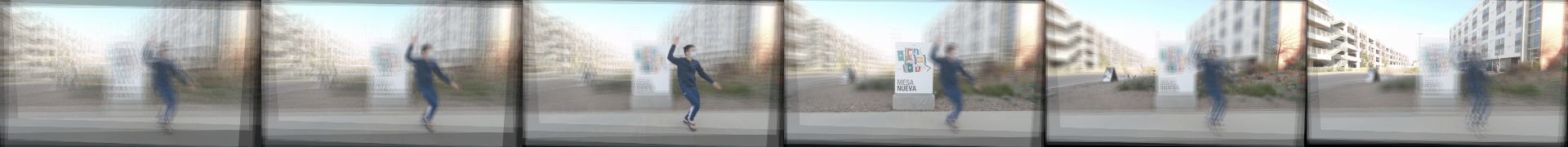}

    \makebox[\textwidth]{\scriptsize Dynamic scene}\vspace{0.05cm}
    \includegraphics[width=\textwidth]{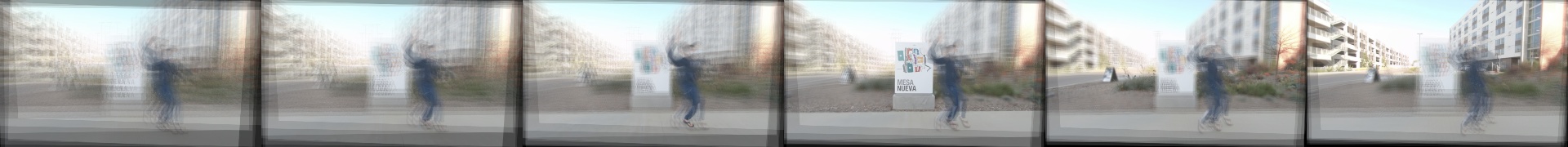}
    
    \caption{\textbf{Dynamic Plane Sweep Volume.} Projecting $V$ input images onto $D$ fronto-parallel planes facing the target camera produces a 5D plane sweep volume tensor of shape $D\!\times\!V\!\times\!3\!\times\!H\!\times\!W$, illustrated here on an example scene by averaging over the $V$ dimension and plotting the $D=6$ planes in a row (the depth increases left-to-right). When the scene is static (top), all the elements of the scene appear in focus at their respective depths. When the scene is dynamic (bottom), static elements are in focus and dynamic elements are not. Our model takes plane sweep volumes as input, computed over dynamic monocular sequences.}
    \label{fig:DPSV}
\end{figure*}

%% file: figures/UCSD.tex
\begin{figure*}[ht]
    \centering
    \makebox[\ucsdb]{\scriptsize Input}\;
    \makebox[\ucsd]{\scriptsize D-3DGS~\cite{yang2024deformable3dgs}}\hfill
    \makebox[\ucsd]{\scriptsize SC-GS~\cite{huang2024sc}}\hfill
    \makebox[\ucsd]{\scriptsize 4DGS~\cite{Wu_2024_CVPR}}\hfill
    \makebox[\ucsd]{\scriptsize GRVS (ours)}\hfill
    \makebox[\ucsd]{\scriptsize Ground-Truth}
    \vspace{0.05cm}

    \begin{minipage}[b]{\ucsdb}
        \includegraphics[width=\textwidth]{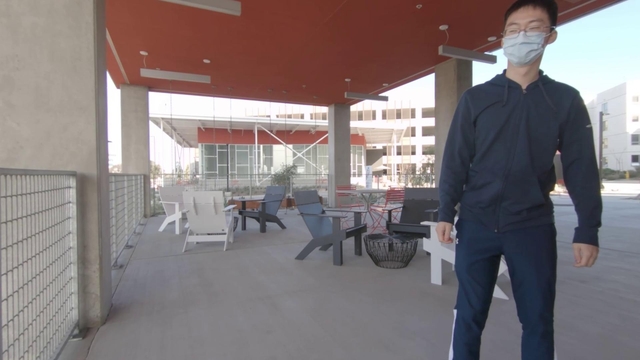}%
        \vspace{-1pt}
        \includegraphics[width=\textwidth]{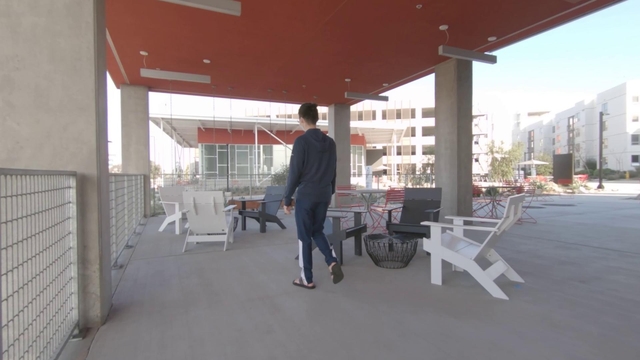}
    \end{minipage}\;
    \includegraphics[width=\ucsd,height=\ucsd]{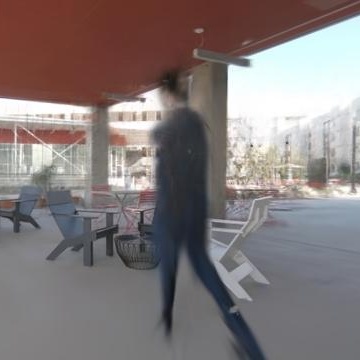}\hfill
    \includegraphics[width=\ucsd,height=\ucsd]{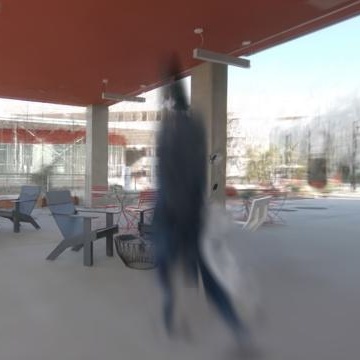}\hfill
    \includegraphics[width=\ucsd,height=\ucsd]{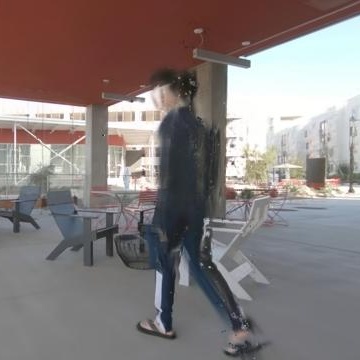}\hfill
    \includegraphics[width=\ucsd,height=\ucsd]{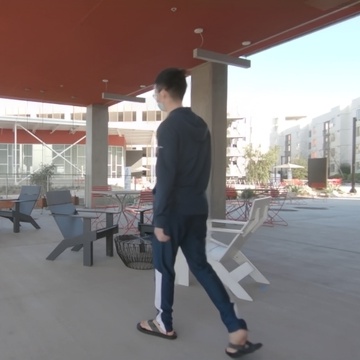}\hfill
    \includegraphics[width=\ucsd,height=\ucsd]{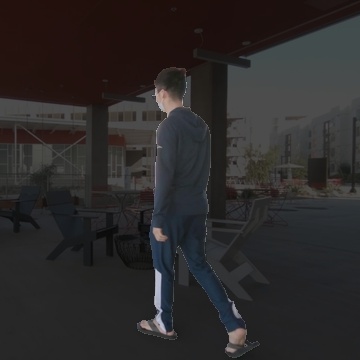}

    \begin{minipage}[b]{\ucsdb}
        \includegraphics[width=\textwidth]{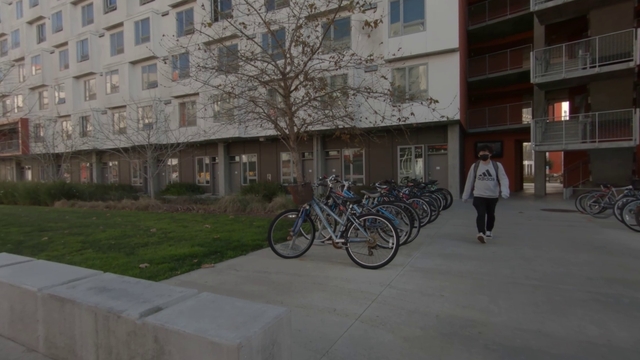}%
        \vspace{-1pt}
        \includegraphics[width=\textwidth]{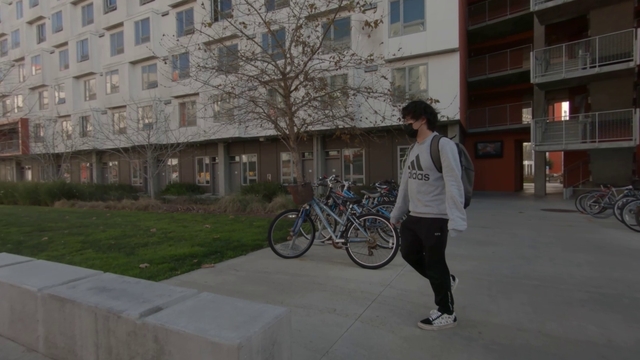}
    \end{minipage}\;
    \includegraphics[width=\ucsd,height=\ucsd]{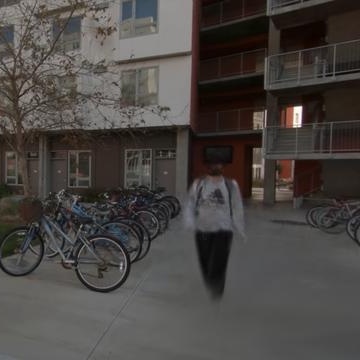}\hfill
    \includegraphics[width=\ucsd,height=\ucsd]{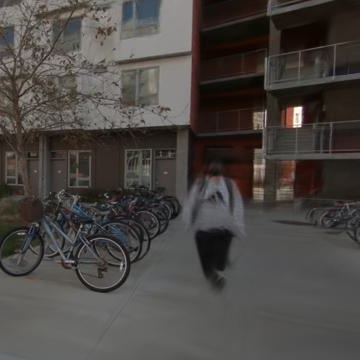}\hfill
    \includegraphics[width=\ucsd,height=\ucsd]{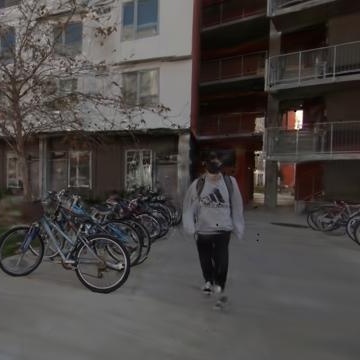}\hfill
    \includegraphics[width=\ucsd,height=\ucsd]{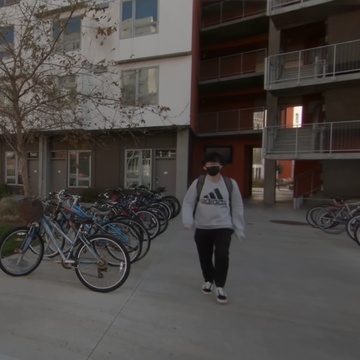}\hfill
    \includegraphics[width=\ucsd,height=\ucsd]{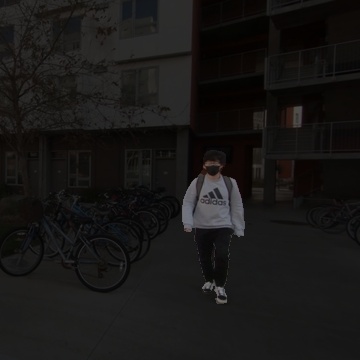}

    \caption{\textbf{Qualitative evaluation on UCSD.} On the left, we show the first and last frames for 2 input sequences. We then show the predictions of the three baselines and our method for a mid-sequence frame on those sequences. The ground-truths are shown on the right, with the dynamic elements highlighted.}
    \label{fig:qualitative-results-UCSD}
\end{figure*}

%% file: figures/Kubric.tex
\begin{figure*}[ht]
    \centering

    \raisebox{-.5\height}{\makebox[0.01\textwidth]{\hspace{-0.2cm} \rotatebox{90}{\scriptsize Input}}}\enspace
    \raisebox{-.5\height}{\includegraphics[width=\kubricb,height=\kubricb]{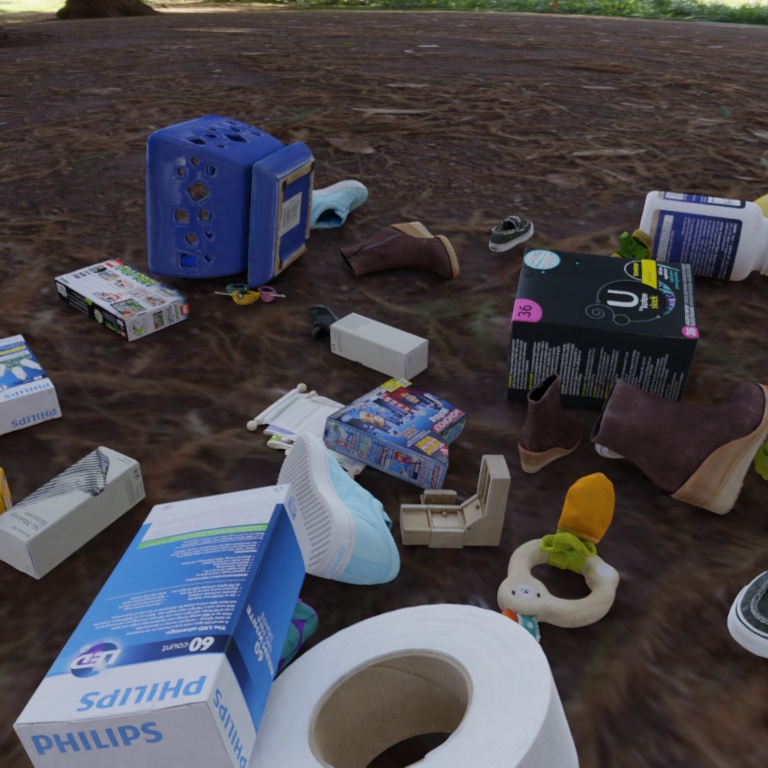}}\hfill
    \raisebox{-.5\height}{\includegraphics[width=\kubricb,height=\kubricb]{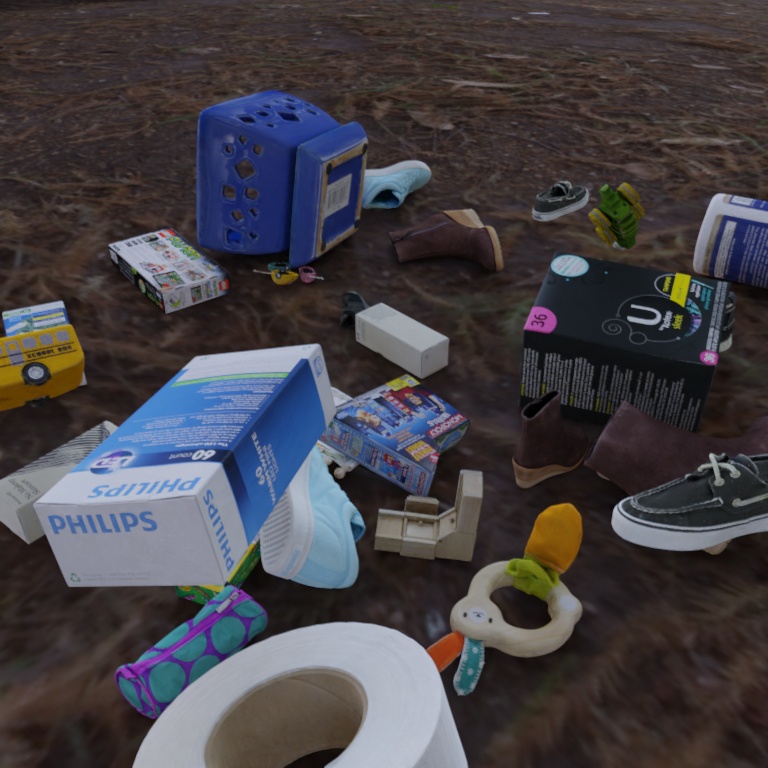}}\hfill
    \raisebox{-.5\height}{\includegraphics[width=\kubricb,height=\kubricb]{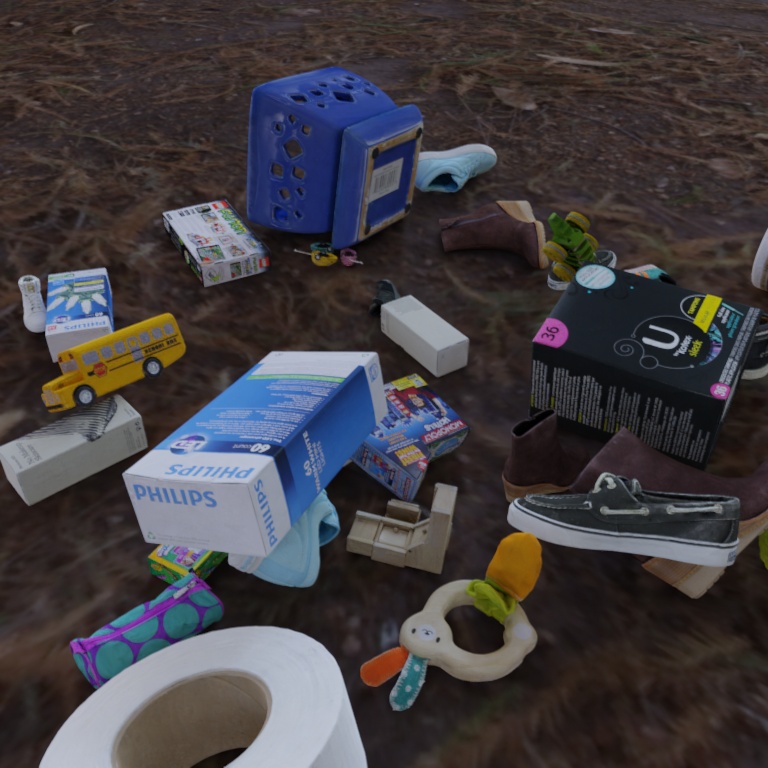}}\hfill
    \raisebox{-.5\height}{\includegraphics[width=\kubricb,height=\kubricb]{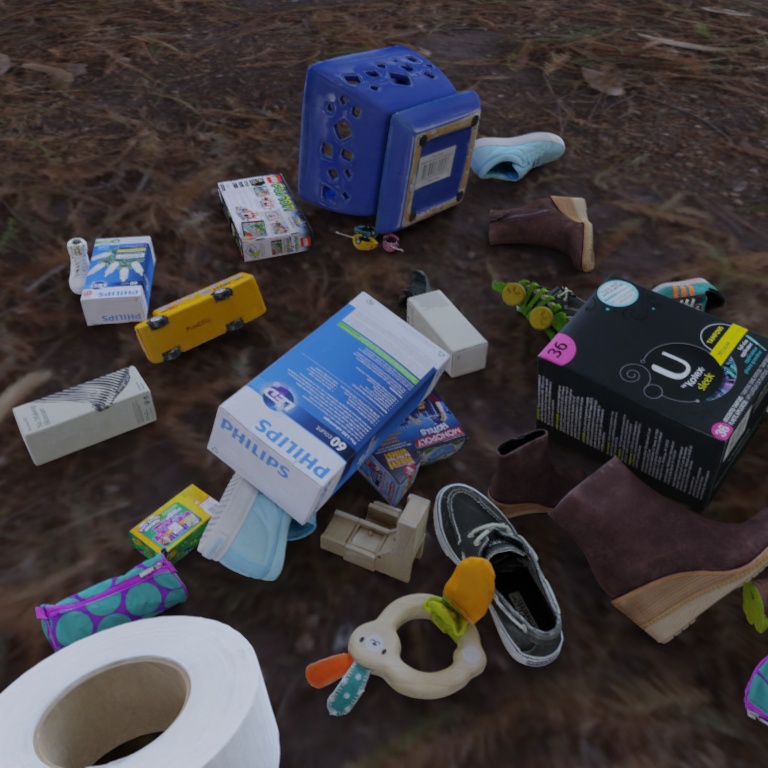}}\hfill
    \raisebox{-.5\height}{\includegraphics[width=\kubricb,height=\kubricb]{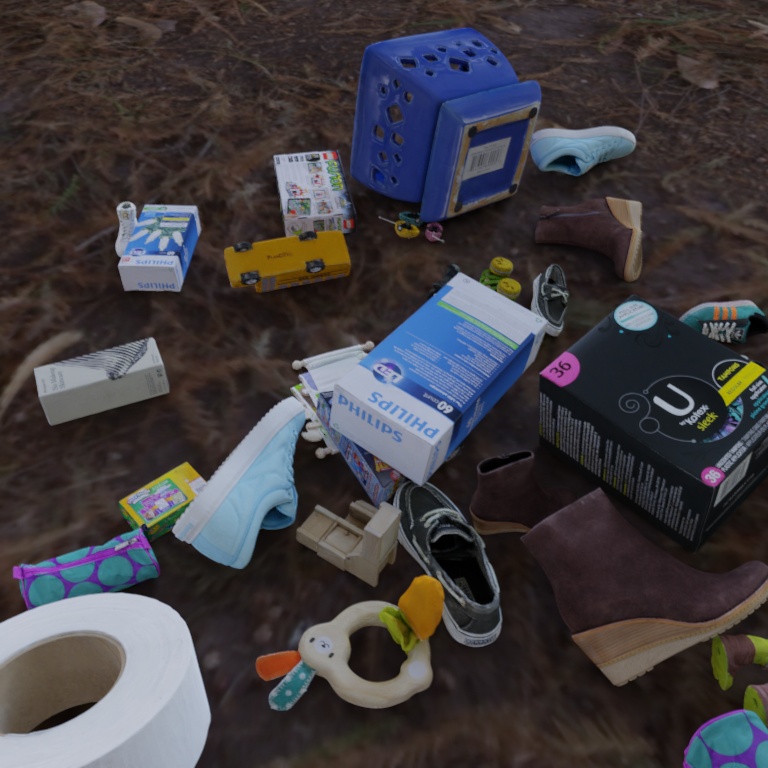}}\hfill
    \raisebox{-.5\height}{\includegraphics[width=\kubricb,height=\kubricb]{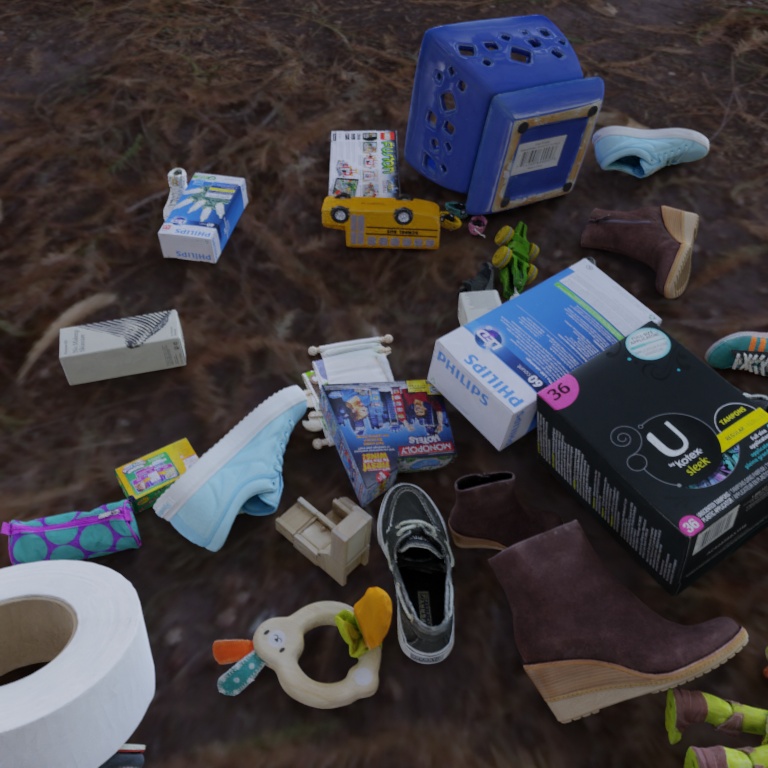}}\hfill
    \raisebox{-.5\height}{\includegraphics[width=\kubricb,height=\kubricb]{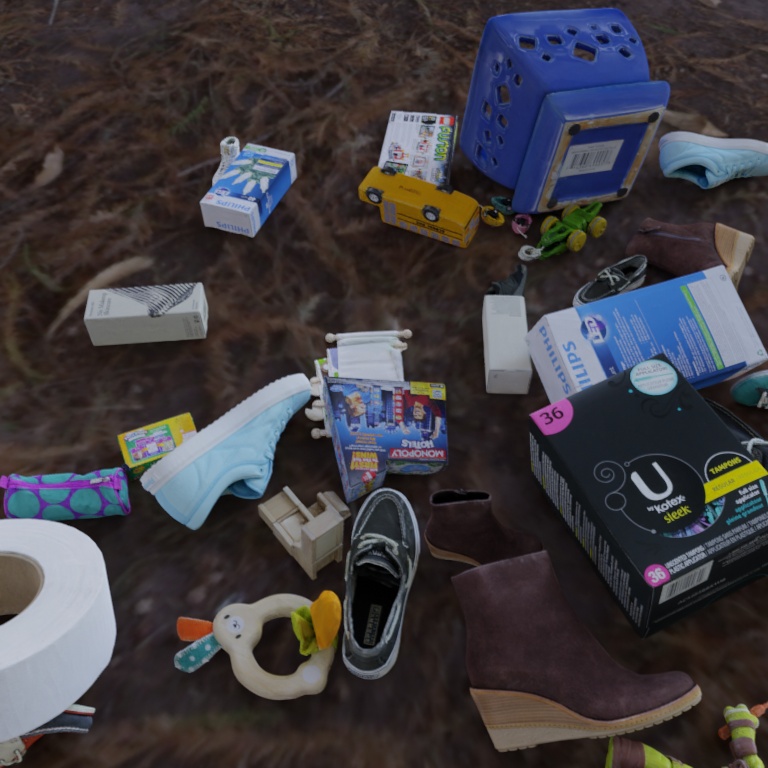}}\hfill
    \raisebox{-.5\height}{\includegraphics[width=\kubricb,height=\kubricb]{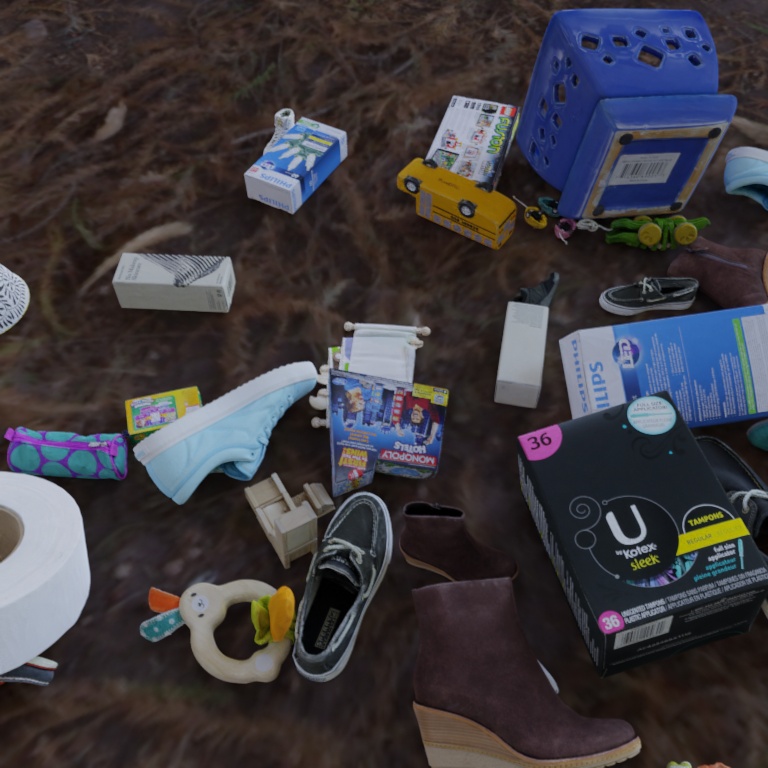}}\hfill
    \raisebox{-.5\height}{\includegraphics[width=\kubricb,height=\kubricb]{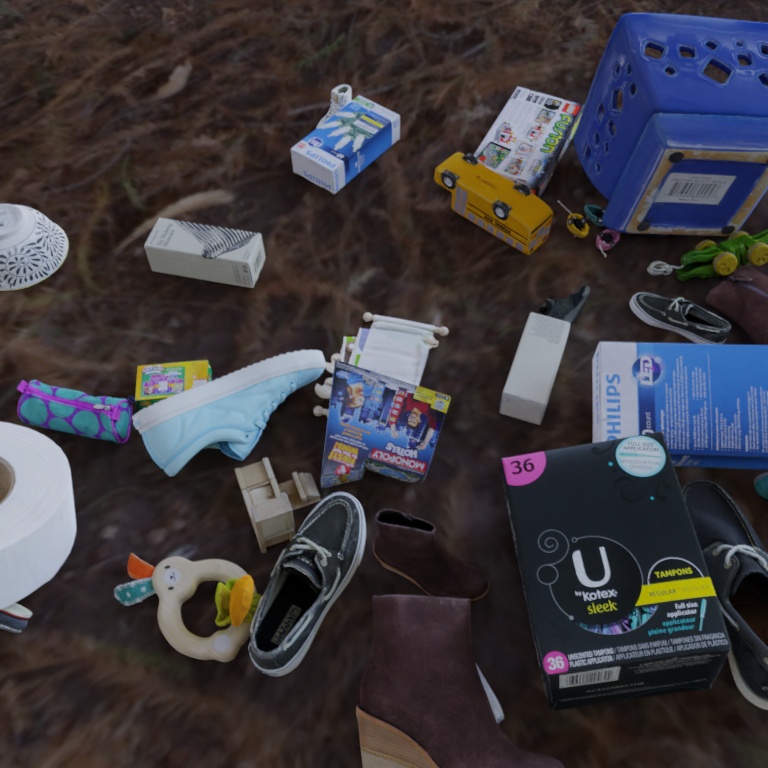}}\hfill
    \raisebox{-.5\height}{\includegraphics[width=\kubricb,height=\kubricb]{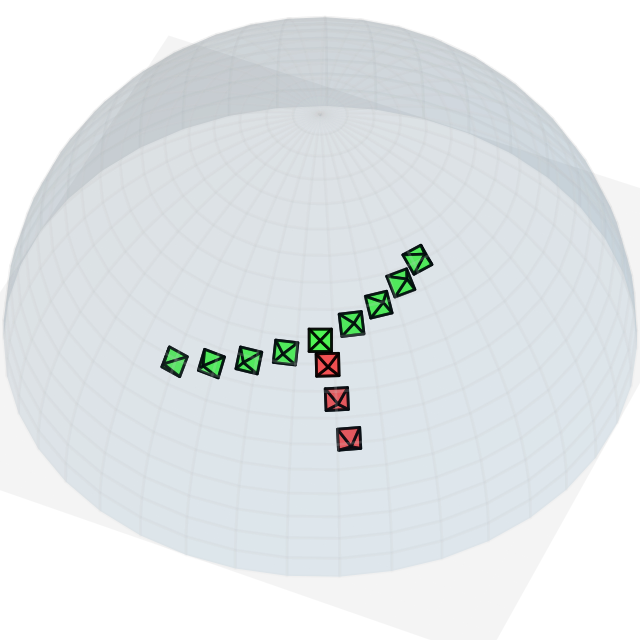}}

    \vspace{0.3cm}
    \raisebox{-.5\height}{\makebox[0.01\textwidth]{\hspace{-0.2cm} \rotatebox{90}{}}}\enspace
    \makebox[\kubric]{\scriptsize D-3DGS~\cite{yang2024deformable3dgs}}\hfill
    \makebox[\kubric]{\scriptsize SC-GS~\cite{huang2024sc}}\hfill
    \makebox[\kubric]{\scriptsize 4DGS~\cite{Wu_2024_CVPR} }\hfill
    \makebox[\kubric]{\scriptsize MoSca~\cite{lei2024mosca} }\hfill
    \makebox[\kubric]{\scriptsize GCD~\cite{vanhoorick2024gcd}}\hfill
    \makebox[\kubric]{\scriptsize Gen3C~\cite{ren2025gen3c}}\hfill
    \makebox[\kubric]{\scriptsize GRVS (ours)}\hfill
    \makebox[\kubric]{\scriptsize GT}
    \vspace{0.05cm}

    \raisebox{-.5\height}{\makebox[0.01\textwidth]{\hspace{-0.2cm} \rotatebox{90}{\scriptsize target 1}}}\enspace
    \raisebox{-.5\height}{\includegraphics[width=\kubric,height=\kubric]{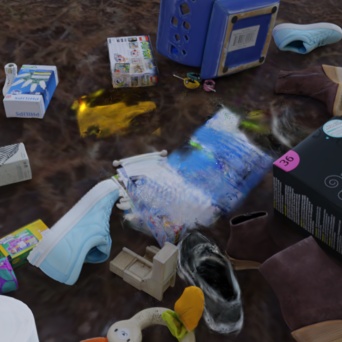}}\hfill
    \raisebox{-.5\height}{\includegraphics[width=\kubric,height=\kubric]{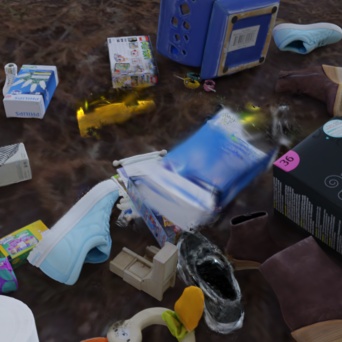}}\hfill
    \raisebox{-.5\height}{\includegraphics[width=\kubric,height=\kubric]{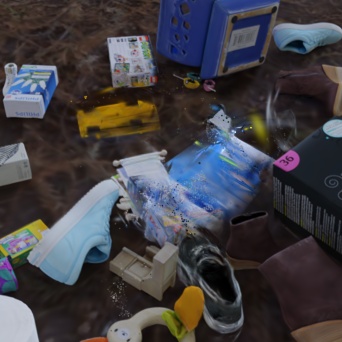}}\hfill
    \raisebox{-.5\height}{\includegraphics[width=\kubric,height=\kubric]{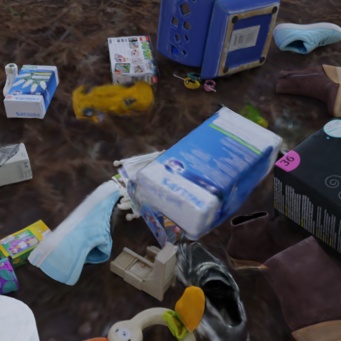}}\hfill
    \raisebox{-.5\height}{\includegraphics[width=\kubric,height=\kubric]{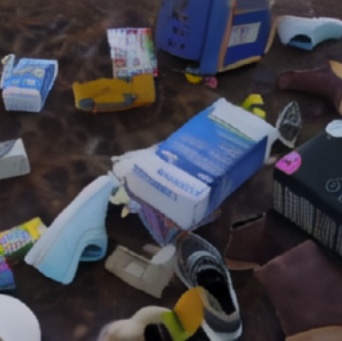}}\hfill
    \raisebox{-.5\height}{\includegraphics[width=\kubric,height=\kubric]{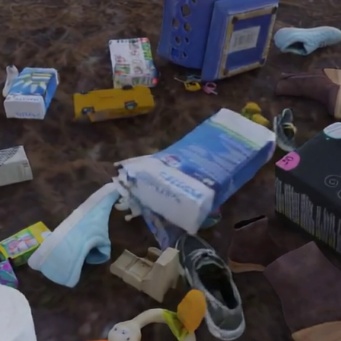}}\hfill
    \raisebox{-.5\height}{\includegraphics[width=\kubric,height=\kubric]{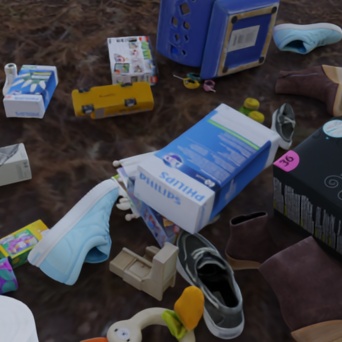}}\hfill
    \raisebox{-.5\height}{\includegraphics[width=\kubric,height=\kubric]{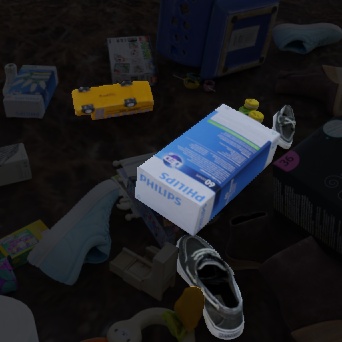}}
    
    \raisebox{-.5\height}{\makebox[0.01\textwidth]{\hspace{-0.2cm} \rotatebox{90}{\scriptsize target 2}}}\enspace
    \raisebox{-.5\height}{\includegraphics[width=\kubric,height=\kubric]{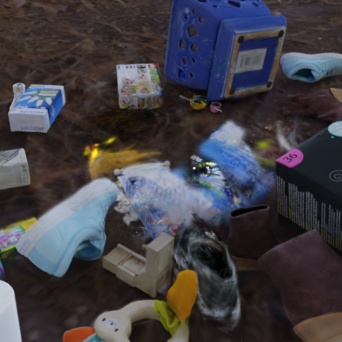}}\hfill
    \raisebox{-.5\height}{\includegraphics[width=\kubric,height=\kubric]{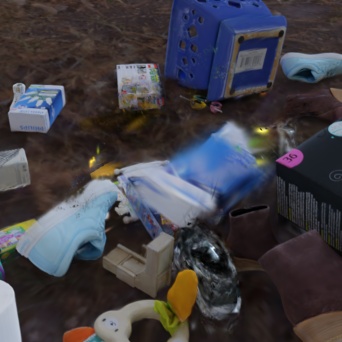}}\hfill
    \raisebox{-.5\height}{\includegraphics[width=\kubric,height=\kubric]{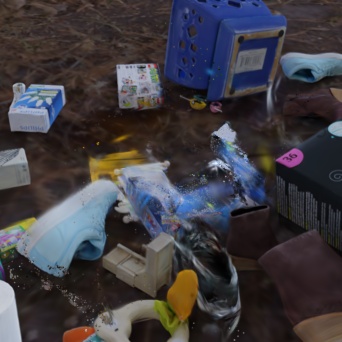}}\hfill
    \raisebox{-.5\height}{\includegraphics[width=\kubric,height=\kubric]{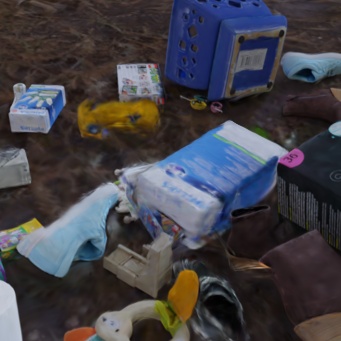}}\hfill
    \raisebox{-.5\height}{\includegraphics[width=\kubric,height=\kubric]{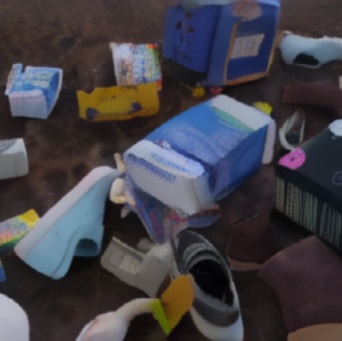}}\hfill
    \raisebox{-.5\height}{\includegraphics[width=\kubric,height=\kubric]{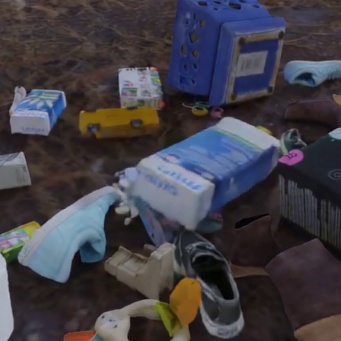}}\hfill
    \raisebox{-.5\height}{\includegraphics[width=\kubric,height=\kubric]{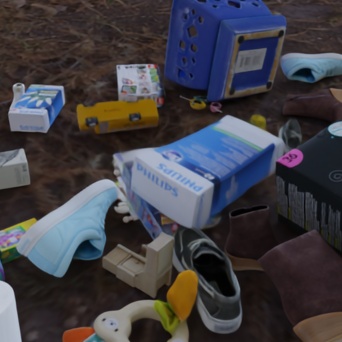}}\hfill
    \raisebox{-.5\height}{\includegraphics[width=\kubric,height=\kubric]{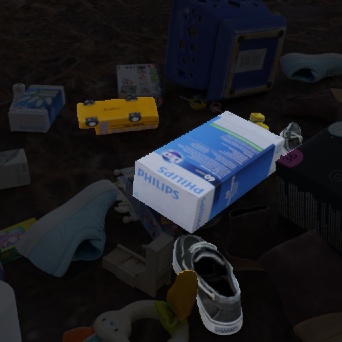}}

    \raisebox{-.5\height}{\makebox[0.01\textwidth]{\hspace{-0.2cm} \rotatebox{90}{\scriptsize target 3}}}\enspace
    \raisebox{-.5\height}{\includegraphics[width=\kubric,height=\kubric]{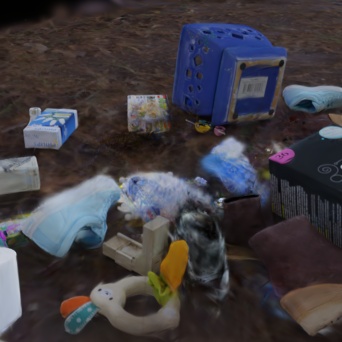}}\hfill
    \raisebox{-.5\height}{\includegraphics[width=\kubric,height=\kubric]{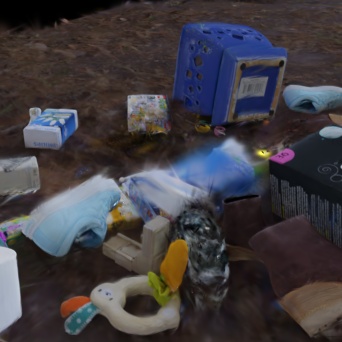}}\hfill
    \raisebox{-.5\height}{\includegraphics[width=\kubric,height=\kubric]{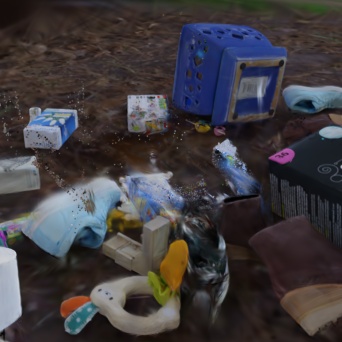}}\hfill
    \raisebox{-.5\height}{\includegraphics[width=\kubric,height=\kubric]{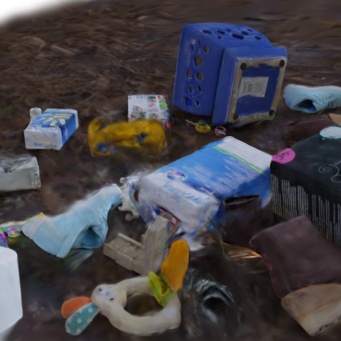}}\hfill
    \raisebox{-.5\height}{\includegraphics[width=\kubric,height=\kubric]{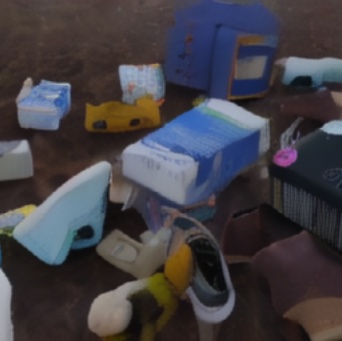}}\hfill
    \raisebox{-.5\height}{\includegraphics[width=\kubric,height=\kubric]{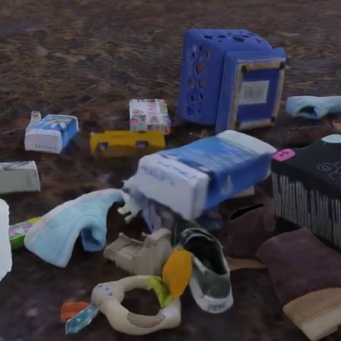}}\hfill
    \raisebox{-.5\height}{\includegraphics[width=\kubric,height=\kubric]{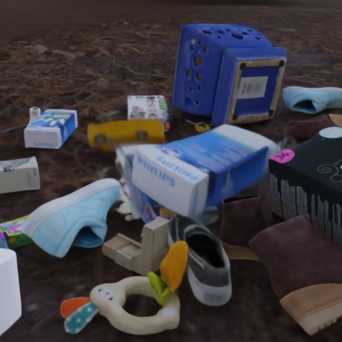}}\hfill
    \raisebox{-.5\height}{\includegraphics[width=\kubric,height=\kubric]{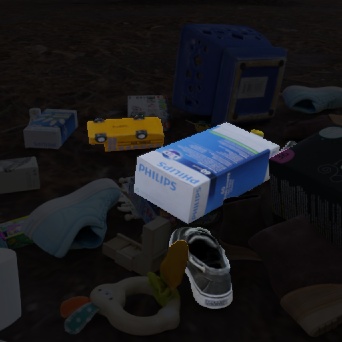}}

    \caption{\textbf{Qualitative evaluation on Kubric-4D-dyn}. At the top, we show every 10\textsuperscript{th} frames for one input sequence, and a visualization of the relative positions of the input trajectory (in green) and three target cameras placed at increasing distances from the input trajectory (in red). Below, we show the predictions of the five baselines and our method for the three target views. The ground-truths are shown on the right, with the dynamic elements highlighted (yellow car, white and blue box, green shoes).}
    \label{fig:qualitative-results-Kubric}
\end{figure*}